\documentclass[onecolumn,12pt]{IEEEtran} 
\usepackage{amsmath,amssymb,epsfig,float} 
\usepackage{times}
\usepackage{algorithm}
\usepackage{amsmath, amssymb}

\newtheorem{theorem}{Theorem}

\newtheorem{lemma}{Lemma}
\newcommand{\qed}{\endIEEEproof}

\newcounter{aid}

\title{Prediction by Random-Walk Perturbation
\thanks{
This research was supported in part by the Natural Sciences and Engineering Research Council (NSERC) of Canada, the Spanish Ministry of Science and Technology grant MTM2012-37195, the Hungarian Scientific Research Fund and the Hungarian National Office for Research and Technology (OTKA-NKTH CNK 77782) and the PASCAL2 Network of Excellence under EC grant no.~216886.}
\thanks{
L.~Devroye is with the School of Computer Science, McGill University, Montreal, Canada H3A 2K6
(email: {\tt lucdevroye@gmail.com}). G.~Lugosi is with ICREA and the Department of Economics, Pompeu Fabra University, Ramon Trias Fargas 25--27, 08005 Barcelona, Spain (email: {\tt gabor.lugosi@gmail.com}). G.~Neu is with the Department of Computer Science and Information Theory,
Budapest University of Technology and Economics, Magyar tud\'osok k\"or\'utja 2, 1117, Budapest, Hungary. He is also with the
Computer and Automation Research Institute of the Hungarian Academy of Sciences, Kende utca 13--17, 1111, Budapest, Hungary (email: \texttt{gergely.neu@gmail.com}).}
}

\author{Luc Devroye, G\'abor Lugosi and Gergely Neu}

\usepackage{bm}
\usepackage{algorithm}
\usepackage[colorinlistoftodos, textwidth=40mm, shadow]{todonotes}

\definecolor{PalePurp}{rgb}{0.66,0.57,0.66}

\newcommand{\loss}{\ell}
\newcommand{\N}{\mathcal{N}}
\newcommand{\I}{\mathcal{I}}
\newcommand{\real}{\mathbb{R}}
\newcommand{\Sw}{\mathcal{S}}

\newcommand{\II}[1]{\mathbb{I}\left\{#1\right\}}
\newcommand{\PP}[1]{\mathbb{P}\left[#1\right]}
\newcommand{\EE}[1]{\mathbb{E}\left[#1\right]}
\newcommand{\PPt}[1]{\mathbb{P}_t\left[#1\right]}
\newcommand{\EEt}[1]{\mathbb{E}_t\left[#1\right]}
\newcommand{\PPc}[2]{\mathbb{P}\left[#1\left|#2\right.\right]}
\newcommand{\PPct}[2]{\mathbb{P}_t\left[#1\left|#2\right.\right]}

\newcommand{\PPcct}[2]{\mathbb{P}_t\left[\left.#1\right|#2\right]}
\newcommand{\EEc}[2]{\mathbb{E}\left[#1\left|#2\right.\right]}

\newcommand{\ev}[1]{\left\{#1\right\}}

\newcommand{\bh}{\boldsymbol{h}}
\newcommand{\bH}{\boldsymbol{H}}
\newcommand{\bX}{\boldsymbol{X}}
\newcommand{\bV}{\boldsymbol{V}}
\newcommand{\bK}{\boldsymbol{K}}
\newcommand{\bw}{\boldsymbol{w}}
\newcommand{\bW}{\boldsymbol{W}}
\newcommand{\bv}{\boldsymbol{v}}
\newcommand{\hV}{\widehat{\bV}}
\newcommand{\hZ}{\widehat{\bZ}}
\newcommand{\bz}{\boldsymbol{z}}
\newcommand{\bloss}{\bm\ell}
\newcommand{\bL}{\boldsymbol{L}}

\newcommand{\bp}{\boldsymbol{p}}
\newcommand{\bZ}{\boldsymbol{Z}}
\newcommand{\PROB}{\mathbb{P}}
\newcommand{\EXP}{\mathbb{E}}
\newcommand{\R}{\mathbb{R}}

\newcommand{\wh}{\widehat}

\begin{document}

\maketitle

\begin{abstract}
We propose a version of the follow-the-perturbed-leader online prediction
algorithm in which the cumulative losses are perturbed by independent
symmetric random walks. The forecaster is shown to achieve an expected regret
of the optimal order $O(\sqrt{n\log N})$ where $n$ is the time horizon
and $N$ is the number of experts. More importantly, it is shown that 
the forecaster changes its prediction at most $O(\sqrt{n\log N})$
times, in expectation. We also extend the analysis to online combinatorial
optimization and show that even in this more general setting, the
forecaster rarely switches between experts while having a regret
of near-optimal order. 
\end{abstract}

\begin{keywords}
Online learning, Online combinatorial optimization, Follow the Perturbed Leader, Random walk
\end{keywords}

\section{Preliminaries}

In this paper we study the problem of online prediction with expert advice, see
\cite{CBLu06:Book}. 
The problem may be described as a repeated game between a {\sl forecaster}
and an adversary---the {\sl environment}. At each time instant $t=1,\ldots,n$, 
the forecaster chooses one of the $N$ available actions 
(often called {\sl experts})
and suffers a loss $\ell_{i,t}\in [0,1]$ corresponding to the chosen action $i$.
We consider the so-called {\sl oblivious adversary} model in which
the environment selects all losses before the prediction game starts
and reveals the losses $\ell_{i,t}$ at time $t$ after the forecaster
has made its prediction. The losses are deterministic but the
forecaster may randomize: at time $t$, the forecaster chooses a
probability distribution $\bp_t$ over the set of $N$ actions
and draws a random action $I_t$ according to the distribution $\bp_t$.
The prediction protocol is described in
Figure~\ref{fig:protocol}. 

The usual goal for the standard prediction
problem is to devise an algorithm such 
that the cumulative loss $\wh{L}_n = \sum_{t=1}^n \loss_{I_t,t}$ is as small 
as possible, in expectation and/or with high probability 
(where probability is with respect
to the forecaster's randomization). Since we do not
make any assumption on how the environment generates the losses
$\loss_{i,t}$, we cannot hope to minimize the above cumulative loss. 
Instead, a meaningful goal is to minimize the performance gap
between our algorithm and the strategy that selects the best action
chosen in hindsight. This performance gap is called the {\sl regret}
and is defined formally as
\[
R_n = \max_{i\in\left\{1,2,\dots,N\right\}} \sum_{t=1}^n\left(\loss_{I_t,t} - \loss_{i,t}\right) = \wh{L}_n - L^*_n,
\]
where  we have also introduced the notation 
$L_n^* =\min_{i\in\left\{1,2,\dots,N\right\}} \sum_{t=1}^n \loss_{i,t}$. 
Minimizing the regret defined above is a well-studied problem. It is known that no matter what algorithm the forecaster
uses,
\[
   \liminf_{n,N\to \infty} \sup \frac{\EXP R_n}{\sqrt{(n/2)\ln N}} \ge 1
\]
where the supremum is taken with respect to all possible
loss assignments with losses in $[0,1]$ (see, e.g., \cite{CBLu06:Book}). 
On the other hand, 
several prediction algorithms are known whose expected regret
is of optimal order $O(\sqrt{n\log N})$ and many of them
achieve a regret of this order with high probability.
Perhaps the most popular one is the exponentially weighted average
forecaster (a variant of weighted majority algorithm of
Littlestone and Warmuth~\cite{LW94}, and aggregating strategies of Vovk~\cite{Vov90}, also
known as {\sl Hedge} by Freund and Schapire~\cite{FrSc97}). The exponentially weighted
average forecaster assigns probabilities to the actions that are 
inversely proportional to an exponential function of the
loss accumulated by each action up to time $t$. 

Another popular forecaster is the {\sl follow the perturbed
  leader} ({\sc fpl}) algorithm of Hannan~\cite{Han57}. 
Kalai and Vempala~\cite{Kalai03efficient} showed that Hannan's forecaster,
when appropriately modified, indeed achieves an expected regret 
of optimal order. At time $t$, 
the {\sc fpl} forecaster adds a random perturbation $Z_{i,t}$
to the cumulative loss $L_{i,t-1}=\sum_{s=1}^{t-1}\loss_{i,s}$
of each action and chooses an action that minimizes the sum 
$L_{i,t-1}+Z_{i,t}$. If the vector of random variables 
$\bZ_t= (Z_{1,t},\ldots,Z_{N,t})$ 
have joint density $(\eta/2)^Ne^{-\eta\|z\|_1}$ for $\eta \sim \sqrt{\log N/n}$,
then the expected regret of the forecaster is of order $O(\sqrt{n\log N})$
(\cite{KV05}, see also \cite{CBLu06:Book}, \cite{HuPo04}, \cite{Pol05}).
This is true whether $\bZ_1,\ldots,\bZ_n$ are independent or not.
It they are independent, then one may show that the regret is concentrated
around its expectation. Another interesting choice is when
$\bZ_1=\cdots=\bZ_n$, that is, the same perturbation is used over time.
Even though this forecaster has an expected regret of optimal order,
the regret is much less concentrated and may fail with reasonably high 
probability.

\begin{figure}
\fbox{
\begin{minipage}{5.9in}
{\bfseries Parameters}: set of actions $\I = \left\{1,2,\dots,N\right\}$, 
number of rounds $n$; \\
The environment chooses the losses $\loss_{i,t}\in [0,1]$ 
for all $i\in\left\{1,2,\dots,N\right\}$ and $t=1,\ldots,n$. \\
{\bfseries For all $t=1,2,\dots,n$, repeat}
\begin{enumerate}
\item The forecaster chooses a probability distribution $\bp_t$ over $\left\{1,2,\dots,N\right\}$.
\item The forecaster draws an action $I_t$ randomly according to $\bp_t$.
\item The environment reveals $\loss_{i,t}$ for all $i\in\left\{1,2,\dots,N\right\}$.
\item The forecaster suffers loss $\loss_{I_t,t}$.
\end{enumerate}
\end{minipage}
}
\caption{Prediction with expert advice.}
\label{fig:protocol}
\end{figure}

Small regret is not the only desirable feature of an online
forecasting algorithm. In many applications, on would like to define
forecasters that do not change their prediction too often.  Examples
of such problems include the online buffering problem described by
Geulen, Voecking and Winkler~\cite{geulen10buffering} and the online lossy source coding problem
of Gy\"orgy and Neu~\cite{gyorgy11nearoptimal}. A more abstract problem where the
number of abrupt switches in the behavior is costly is the problem of
online learning in Markovian decision processes, as described by
Even-Dar, Kakade and Mansour~\cite{even-dar09OnlineMDP} and Neu, Gy\"orgy, Szepesv\'ari and Antos~\cite{neu10o-mdp}.

To be precise, define the number of action switches up to time $n$ by
\[
C_n = \left|\left\{1<t\le n:I_{t-1}\neq I_t\right\}\right|~.
\]
In particular, we are interested in defining randomized forecasters 
that achieve a regret $R_n$ of the order $O(\sqrt{n\log N})$ while
keeping the number of action switches $C_n$ as small as possible.
However, the usual forecasters with small regret---such as the 
exponentially weighted average forecaster or the {\sc fpl} forecaster
with i.i.d.\ perturbations---may switch actions a large number of times, typically $\Theta(n)$. Therefore, the design of special forecasters 
with small regret and small number of action switches is called for.

The first paper
to explicitly attack this problem is by Geulen, Voecking and Winkler~\cite{geulen10buffering}, who
propose a variant of the exponentially weighted average forecaster
called the ``Shrinking Dartboard'' algorithm and prove that it
provides an expected regret of $O(\sqrt{n\log N})$, while
guaranteeing that the 
expected number of switches is at most $O(\sqrt{n\log N})$. 
A less conscious attempt to solve the
problem is due to Kalai and Vempala~\cite{KV05}; they show
that the 
simplified version of the {\sc fpl} algorithm with identical perturbations
(as described above) guarantees an
$O(\sqrt{n\log N})$ bound on both the expected regret and the
expected number of switches. 
In this paper, we
propose a method based on {\sc fpl} in which perturbations are defined
by independent symmetric random walks. We show that this, intuitively
appealing, forecaster has similar regret and switch-number guarantees
as Shrinking Dartboard and {\sc fpl} with identical perturbations.  A
further important advantage of the new forecaster is that it may be
used simply in the more general problem of {\sl online combinatorial}---or, 
more generally, {\sl linear}---{\sl optimization}.
We postpone the definitions and the statement of the results
to Section \ref{sec:comb} below.

\section{The algorithm}

To address the problem described in the previous section, we propose a
variant of the Follow the Perturbed Leader ({\sc fpl}) algorithm.
The proposed forecaster perturbs the loss of each action at every time
instant by a symmetric coin flip
and chooses an action with minimal cumulative perturbed loss.
More precisely, the algorithm draws  independent random variables 
$X_{i,t}$ that take values $\pm 1/2$ with equal probabilities and
$X_{i,t}$ is added to each loss
$\loss_{i,t-1}$. At time $t$ action $i$ is chosen that minimizes
$\sum_{s=1}^t \left(\loss_{i,t-1} + X_{i,t}\right)$ 
(where we define $\loss_{i,0}=0$).

\begin{algorithm}
{\bfseries Initialization}: set $L_{i,0}=0$ and $Z_{i,0}=0$ for all $i=1,2,\dots,N$.\\
{\bfseries For all $t=1,2,\dots,n$, repeat}
\begin{enumerate}
\item Draw $X_{i,t}$ for all $i=1,2,\dots,N$ such that
\[
X_{i,t} = 
\begin{cases}
\frac 12 &\mbox{with probability $\frac 12$}\\
-\frac 12 &\mbox{with probability $\frac 12$.}
\end{cases}
\]
\item Let $Z_{i,t}=Z_{i,t-1} + X_{i,t}$ for all $i=1,2,\dots,N$.
\item Choose action 
\[
I_t = \arg\min_i \left(L_{i,t-1} + Z_{i,t}\right).
\]
\item Observe losses $\loss_{i,t}$ for all $i=1,2,\dots,N$, suffer loss $\loss_{I_t,t}$.
\item Set $L_{i,t} = L_{i,t-1} + \loss_{i,t}$ for all $i=1,2,\dots,N$.
\end{enumerate}
\caption{Prediction by random-walk perturbation.}
\label{alg:fppl}
\end{algorithm}

Equivalently, the forecaster may be thought of as an {\sc fpl}
algorithm in which the cumulative losses $L_{i,t-1}$ are perturbed
by $Z_{i,t}=\sum_{i=1}^t X_{i,t}$. Since for each fixed $i$, 
$Z_{i,1},Z_{i,2},\ldots$ is a symmetric random walk, cumulative losses 
of the $N$ actions are perturbed by $N$ independent symmetric random walks.
This is the way the algorithm
is presented in Algorithm~\ref{alg:fppl}. 

A simple variation is when one replaces random coin flips by 
independent standard normal random variables. 
Both have similar performance guarantees and we choose $\pm (1/2)$-valued
perturbations for mathematical convenience. In Section \ref{sec:comb}
we switch to normally distributed perturbations---again driven by mathematical
simplicity. In practice both versions are expected to have a similar
behavior.

Conceptually, the difference between standard {\sc fpl} and 
the proposed version is
the way the perturbations are generated: while common versions of 
{\sc fpl} use perturbations that are generated in an i.i.d.\ fashion, the
perturbations of the algorithm proposed here are dependent.  
This will enable us to
control the number of action switches during the learning
process. Note that the standard deviation of these perturbations is
still of order $\sqrt{t}$ just like for the standard {\sc fpl} 
forecaster with optimal parameter settings.

To obtain intuition why this approach will solve our problem, first
consider a problem with $N=2$ actions and an environment that generates
equal losses, say $\ell_{i,t}=0$ for all $i$ and $t$, 
for all actions. When using i.i.d.\ perturbations, 
{\sc fpl} switches actions with probability $1/2$ in each round, thus
yielding $C_t = t/2 + O(\sqrt{t})$ with overwhelming probability. 
The same holds for the exponentially weighted average forecaster.
On
the other hand, when using the random-walk perturbations described
above, we only switch between the actions when the leading random walk
is changed, that is, when the difference of the two random walks---which
is also a symmetric random walk---hits
zero. It is a well known that the number of occurrences of this
event up to time $t$ is $O_p(\sqrt{t})$, see, \cite{Fel68}. 
As we show below, this is the worst case for the number of switches.

\section{Performance bounds}

The next theorem summarizes our performance bounds for the proposed
forecaster.

\begin{theorem}
\label{th:main}
The expected regret and expected number of switches of actions of
the forecaster of Algorithm~\ref{alg:fppl} satisfy, for all
possible loss sequences (under the oblivious-adversary model),
\[
 \EXP R_n \le 2\EXP C_n \le 8\sqrt{2n\log N} + 16\log n + 16~.
\]
\end{theorem}

\noindent
{\bf Remark.}
Even though we only prove bounds for the expected regret and 
the expected number of switches, it is of great interest to understand
upper tail probabilities. However, this is a highly nontrivial
problem. One may get an intuition by considering the case
when $N=2$ and all losses are equal to zero. In this case 
the algorithm switches actions whenever a symmetric random walk
returns to zero. This distribution is well understood and
the probability that this occurs more than $x\sqrt{n}$ times
during the first $n$ steps is roughly $2\PROB\{N> 2x\}\le 2e^{-2x^2}$
where $N$ is a standard normal random variable
(see \cite[Section III.4]{Fel68}). Thus, in this case we see that
the number of switches is bounded by
$O\left(\sqrt{n\log (1/\delta)}\right)$, with probability at least 
$1-\delta$. However, proving analog bounds for the general case remains
a challenge.

To prove the theorem, we first show that the 
regret can be bounded in terms of the number of action
switches. Then we turn to analyzing the expected number of action
switches.

\subsection{Regret and number of switches}

The next simple lemma shows that the regret of the forecaster may be bounded in terms of the number of times the forecaster switches actions.

\begin{lemma}\label{lem:experts}
Fix any $i\in\left\{1,2,\dots,N\right\}$. Then
\[
\wh{L}_n - L_{i,n} \le 2C_n + Z_{i,n+1} - \sum_{t=1}^{n+1} X_{I_{t-1},t}~.
\]
\end{lemma}
\begin{IEEEproof}
We apply Lemma~3.1 of \cite{CBLu06:Book} (sometimes referred to as the
{\sl ``be-the-leader''} lemma)
for the sequence $(\loss_{\cdot,t-1}+X_{\cdot,t})_{t=1}^\infty$ with $\loss_{j,0}=0$ for all $j\in\left\{1,2,\dots,N\right\}$, obtaining
\[
\begin{split}
\sum_{t=1}^{n+1} \left(\loss_{I_t,t-1} + X_{I_t,t}\right) 
&\le \sum_{t=1}^{n+1} \left(\loss_{i,t-1} + X_{i,t}\right)
\\
&= L_{i,n} + Z_{i,n+1}~. 
\end{split} 
\]
Reordering terms, we get
\begin{equation}\label{eq:fpl_int}
\sum_{t=1}^n \loss_{I_t,t} \le L_{i,n} + \sum_{t=1}^{n+1} \left(\loss_{I_{t-1},t-1} - \loss_{I_{t},t-1}\right) + Z_{i,n} - \sum_{t=1}^{n+1} X_{I_t,t}~.
\end{equation}
The last term can be rewritten as
\[
- \sum_{t=1}^{n+1} X_{I_t,t} = -\sum_{t=1}^{n+1} X_{I_{t-1},t} + \sum_{t=1}^{n+1} \left(X_{I_{t-1},t} - X_{I_{t},t}\right)~.
\]
Now notice that $X_{I_{t-1},t} - X_{I_{t},t}$ and $\loss_{I_{t-1},t-1} - \loss_{I_{t},t-1}$ are both zero when $I_t=I_{t-1}$ and are upper bounded by $1$ otherwise. That is, we get that
\[
\sum_{t=1}^{n+1} \left(\loss_{I_{t-1},t-1} - \loss_{I_{t},t-1}\right) + \sum_{t=1}^{n+1} \left(X_{I_{t-1},t} - X_{I_t,t}\right) \le 2\sum_{t=1}^{n+1} \II{I_{t-1}\neq I_t} = 2C_n~.
\]
Putting everything together gives the statement of the lemma.
\end{IEEEproof}



\subsection{Bounding the number of switches}

Next we analyze the number of switches $C_n$. In particular, we upper bound the marginal probability
$\PP{I_{t+1}\neq I_t}$ for each $t\ge 1$.  
We define the {\sl lead pack} $A_t$ as the set of actions that, at time $t$, have a positive probability of taking the lead at time $t+1$:
\[
A_t = \left\{i\in\left\{1,2,\dots,N\right\}: L_{i,t-1} + Z_{i,t} \le \min_j \left(L_{j,t-1} + Z_{j,t}\right) + 2\right\}~.
\]
We bound the probability of lead change as
\[
\PP{I_t\neq I_{t+1}} \le 
 \frac 12 \PP{|A_t|>1}~.
\]
The key to the proof of the theorem is the
following lemma that gives an upper bound for the probability 
that the lead pack contains more than one action. 
It implies, in particular, that
\[
\EE{C_n} \le 4\sqrt{2n\log N} + 4\log n + 4~,
\]
which is what we need to prove the expected-value bounds of Theorem \ref{th:main}.

\begin{lemma}\label{lem:leadpack}
\[
\PP{|A_t|>1} \le 4\sqrt{2\frac{\log N}{t}} + \frac {8}{t}~.
\]
\end{lemma}

\begin{IEEEproof}
Define $p_t(k) = \PP{Z_{i,t} = \frac k2}$ for all $k=-t,\dots,t$ and
we let $S_t$ denote the set of leaders at time $t$ (so that the forecaster
picks $I_t\in S_t$ arbitrarily): 
\[
S_t = \left\{j\in\left\{1,2,\dots,N\right\}: L_{j,t-1} + Z_{j,t} = \min_i \left\{L_{i,t-1} + Z_{i,t}\right\}\right\}~.
\]
Let us start with analyzing $\PP{|A_t|=1}$:
\[
\begin{split}
\PP{|A_t|=1} =& \sum_{k=-t}^{t} \sum_{j=1}^N p_{t}(k) \PP{\min_{i \in \left\{1,2,\dots,N\right\}\setminus j} \left\{L_{i,t-1} + Z_{i,t}\right\} \ge L_{j,t-1} + \frac{k}{2} + 2}
\\
\ge& \sum_{k=-t}^{t-4} \sum_{j=1}^N p_{t}(k+4) \PP{\min_{i \in \left\{1,2,\dots,N\right\}\setminus j} \left\{L_{i,t-1} + Z_{i,t}\right\} \ge L_{j,t-1} + \frac{k+4}{2}} \frac{p_{t}(k)}{p_{t}(k+4)}
\\
=& \sum_{k=-t+4}^{t} \sum_{j=1}^N p_{t}(k) \PP{\min_{i \in \left\{1,2,\dots,N\right\}\setminus j} \left\{L_{i,t-1} + Z_{i,t}\right\} \ge L_{j,t-1} + \frac k2} \frac{p_{t}(k-4)}{p_{t}(k)}~.
\end{split}
\]
Before proceeding, we need to make two observations. First of all,
\[
\begin{split}
\sum_{j=1}^N p_{t}(k) \PP{\min_{i\in\left\{1,2,\dots,N\right\}\setminus j} \left\{L_{i,t-1} + Z_{i,t}\right\} \ge L_{j,t-1} +  \frac k2} &\ge \PP{\exists j\in S_t: Z_{j,t} = \frac k2} 
\\
&\ge \PP{\min_{j\in S_t} Z_{j,t} = \frac k2},
\end{split}
\]
where the first inequality follows from the union bound and the second from the fact that the latter event implies the former.
Also notice that 
$Z_{i,t} + \frac{t}{2}$ is binomially distributed with parameters
$t$ and $1/2$
and therefore $p_{t}(k) = {\binom{t}{\frac{t+k}{2}}} \frac{1}{2^t}$.
Hence
\[
\begin{split}
\frac{p_{t}(k-4)}{p_{t}(k)} &= \frac{\left(\frac{t+k}{2}\right)! \left(\frac{t-k}{2}\right)!}{\left(\frac{t+k}{2}-2\right)! \left(\frac{t-k}{2}+2\right)!}
\\
&= 1 + \frac{4(t+1)(k - 2)}{(t-k+2) (t-k+4)}~.
\end{split}
\]
It can be easily verified that
\[
\frac{4(t+1)(k - 2)}{(t-k+2) (t-k+4)} \ge \frac{4(t+1)(k - 2)}{(t+2) (t+4)}
\]
holds for all $k\in\left[-t, t\right]$.
Using our first observation, we get
\[
\begin{split}
\PP{|A_t|=1}
\ge& \sum_j \sum_{k=-t+4}^t p_t(k) \PP{\min_{i\in\left\{1,2,\dots,N\right\}\setminus j} \left\{L_{i,t-1}+ Z_{l,t}\right\} \ge L_{j,t-1} + \frac k2} \frac{p_{t}(k-4)}{p_{t}(k)}
\\
\ge& \sum_{k=-t+4}^t \PP{\min_{j\in S_t} Z_{j,t} = \frac{k}{2}} \frac{p_{t}(k-4)}{p_{t}(k)}~.
\end{split}
\]
Along with our second observation, this implies
\[
\begin{split}
\PP{|A_t|>1}
\le& 
1-\sum_{k=-t+4}^{t} \PP{\min_{j\in S_t} Z_{j,t} = \frac k2} \frac{p_{t}(k-4)}{p_{t}(k)}
\\
\le& 
1-\sum_{k=-t+4}^{t} \PP{\min_{j\in S_t} Z_{j,t} = \frac k2} \left(1 + \frac{4(t+1)(k - 2)}{(t+2) (t+4)}\right)
\\
\le& 
\sum_{k=-t}^{t} \PP{\min_{j\in S_t} Z_{j,t} = \frac k2} \left(\frac{4(2-k)(t + 1)}{(t+2) (t+4)}\right)
\\
=& 
\frac{8(t+1)}{(t+2)(t+4)} - 8\frac{t+1}{(t+2)(t+4)} \EE{\min_{j\in S_t} Z_{j,t}}
\\
\le& \frac{8}{t} + \frac{8}{t} \EE{\max_{j\in \left\{1,2,\dots,N\right\}} Z_{j,t}}~.
\end{split}
\]
Now using $\EE{\max_j Z_{j,t}} \le \sqrt{\frac{t\log N}{2}}$
implies
\[
\PP{|A_t|>1} \le  4\sqrt{\frac{2\log N}{t}} + \frac{8}{t}
\]
as desired.
\end{IEEEproof}

\section{Online combinatorial optimization}
\label{sec:comb}

In this section we study the case of online linear optimization
(see, among others, 
\cite{GW98},
\cite{KW01}, 
\cite{GLS01}, 
\cite{TW03}, 
\cite{KV05}, 
\cite{WK08}, 
\cite{HW09},
\cite{HKW10},
\cite{KWK10},
\cite{CL12},
\cite{audibert2011minimax}).
This is a similar prediction problem as the one described in the 
introduction but here each action $i$ is represented by a
vector $\bv_i\in\R^d$. The loss corresponding to action $i$ at time
$t$ equals $\bv_i^\top \bloss_t$
where $\bloss_t \in [0,1]^d$ is the so-called {\sl loss vector}.
Thus, given a set of actions
$\Sw=\ev{\bv_i:i=1,2,\dots,N}\subseteq \R^d$,
at every time instant $t$, the forecaster chooses, in a possibly 
randomized way, a vector $\bV_t \in \Sw$ and suffers loss 
$\bV_t^\top \bloss_t$. We denote by $\wh{L}_n= \sum_{t=1}^n \bV_t^\top \bloss_t$
the cumulative loss of the forecaster and 
the regret becomes
\[
   \wh{L}_n - \min_{\bv\in\Sw} \bv^\top \bL_n 
\]
where $\bL_t = \sum_{s=1}^t \bloss_s$ is the cumulative loss vector.
While the results of the previous section still hold when treating each $\bv_i\in \Sw$ as a separate action,
 one may gain important 
computational advantage by taking the structure of the action set
into account. In particular, as \cite{KV05} emphasize, {\sc fpl}-type
forecasters may often be computed efficiently. In this section 
we propose such a forecaster which adds independent random-walk perturbations
to the {\sl individual components} of the loss vector.
To gain simplicity in the presentation, we restrict our
attention to the case of {\sl online combinatorial optimization}  
in which 
$\Sw \subset \ev{0,1}^d$, that is, each action is represented a binary vector.
This special case arguably contains most important applications 
such as the {\sl online shortest path} problem. In this example,
a fixed directed acyclic graph of $d$ edges is given with two distinguished 
vertices $u$ and $w$. The forecaster, at every time instant $t$, 
chooses a directed path from $u$ to $w$. Such a path is represented
by it binary incidence vector  $\bv\in \ev{0,1}^d$. The components
of the loss vector $\bloss_t\in [0,1]^d$ represent losses assigned to 
the $d$ edges and $\bv^\top \bloss_t$ is the total loss assigned to the
path $\bv$. 
Another (non-essential) simplifying assumption is that every action
$\bv\in \Sw$ has the same number of $1$'s: $\|\bv\|_1=m$ for all $\bv\in\Sw$.
The value of $m$ plays an important role in the bounds below.

The proposed prediction algorithm is defined as follows.
Let $\bX_1,\ldots,\bX_n$ be independent Gaussian random vectors
taking values in $\R^d$ such that the components of each $\bX_t$ 
are i.i.d.\ normal 
$X_{i,t} \sim \N(0,\eta^2)$ 
for some 
fixed $\eta>0$ whose value will be specified later.
Denote
\[
\bZ_{t} = \sum_{s=1}^t \bX_{t}~.
\]
The forecaster at time $t$, chooses the action
\[
\bV_t = \arg\min_{\bv\in\Sw} \ev{\bv^\top \left(\bL_{t-1} + \bZ_t\right)},
\]
where $\bL_t = \sum_{s=1}^t \bloss_t$ for $t\ge 1$ and $\bL_0=(0,\ldots,0)^\top$.

The next theorem bounds the performance of the proposed forecaster.
Again, we are not only interested in the regret but also
the number of switches $\sum_{t=1}^n \II{\bV_{t+1}\neq \bV_t}$.
The regret is of similar order---roughly $m\sqrt{dn}$---as that of the 
standard {\sc fpl} forecaster, up to a logarithmic factor.
Moreover, the expected number of switches is
$O\left(m^2(\log d)^{5/2}\sqrt{n}\right)$. Remarkably, the dependence on $d$
is only polylogarithmic and it is the weight $m$ of the actions
that plays an important role. 

We note in passing that the Shrinking Dartboard algorithm of \cite{geulen10buffering} can be used for simultaneously guaranteeing that the expected regret is $O(m^{3/2}\sqrt{n\log d})$ and the expected number of switches is $\sqrt{mn\log d}$. However, as this algorithm requires explicit computation of the exponential weighted forecaster, it can only be efficiently implemented for some special decision sets $\Sw$---see \cite{KWK10} and \cite{CL12} for some examples. On the other hand, our algorithm can be efficiently implemented whenever there exists an efficient implementation of the static optimization problem of finding $\arg\min_{\bv\in\Sw} \bv^\top \bloss$ for any $\bloss\in\real^d$.

\begin{theorem}\label{thm:combbound}
Fix any $\bv\in\Sw$. The expected regret and the expected number of action switches
satisfy (under the oblivious adversary model)
\[
\EXP \wh{L}_n - \bv^\top \bL_{n} 
\le m\sqrt{n}\left(\frac{2d}{\eta} + \eta \sqrt{2\log d}\right) + \frac{md(\log n + 1)}{\eta^2}
\]
and
\[
\begin{split}\EXP \sum_{t=1}^n \II{\bV_{t+1}\neq \bV_t}
\le& 
\sum_{t=1}^n \frac{m\left(1+2\eta\left(2\log d +\sqrt{2\log d}+1\right)
 + \eta^2\left(2\log d +\sqrt{2\log d}+1\right)^2\right)}{4\eta^2t} 
\\
&  + \sum_{t=1}^n
\frac{m\left(1+\eta\left(2\log d +\sqrt{2\log d}+1\right)\right)\sqrt{2\log d}}{\eta\sqrt{t}}.
\end{split}
\]
In particular, setting $\eta = \sqrt{\frac{2d}{\sqrt{2\log d}}}$ yields
\[
\EXP\wh{L}_n - \bv^\top \bL_{n} 
\le 4m\sqrt{dn}\sqrt[4]{\log d} + m(\log n + 1)\sqrt{\log d}.
\]
and
\[
\begin{split}
\EXP \sum_{t=1}^n \II{\bV_{t+1}\neq \bV_t} = O\left(m(\log d)^{5/2}\sqrt{n}\right).
\end{split}
\]
\end{theorem}

The proof of the regret bound is quite standard, similar to the proof of Theorem~3 in
\cite{audibert12regret}, and is deferred to the appendix. 
The more interesting part
is the bound for the expected number of action switches
$\EXP \sum_{t=1}^n \II{\bV_{t+1}\neq \bV_t}=
\sum_{t=1}^n \PP{\bV_{t+1}\neq \bV_t}$. It follows from the
lemma below and the well-known fact that the expected value
of the maximum of the square of $d$ independent standard normal 
random variables is at most $2\log d +\sqrt{2\log d}+1$
(see, e.g., \cite{BoLuMa13}). Thus, it suffices to prove the following:

\begin{lemma}
For each $t=1,2,\ldots,n$,
\[
\PPc{\bV_{t+1}\neq \bV_t}{\bX_{t+1}} \le \frac{m\left\|\bloss_t + \bX_{t+1}\right\|_\infty^2}{2\eta^2t} + \frac{m\left\|\bloss_t + \bX_{t+1}\right\|_\infty\sqrt{2\log d}}{\eta\sqrt{t}}
\]
\end{lemma}

\begin{IEEEproof}
We use the notation $\PPt{\cdot} = \PPc{\cdot}{\bX_{t+1}}$ and $\EEt{\cdot} = \EEc{\cdot}{\bX_{t+1}}$. Also, let
\[
\bh_t = \bloss_t + \bX_{t+1}\qquad\qquad\mbox{and}\qquad\qquad \bH_t = \sum_{s=0}^{t-1} \bh_t.
\]
Furthermore, we will use the shorthand notation $c = \left\|\bh_t\right\|_\infty$.
Define the set $A_t$ as the lead pack:
\[
A_t = \ev{\bw\in\Sw: (\bw-\bV_t)^\top \bH_t \le \left\|\bw-\bV_t\right\|_1c}~.
\]
Observe that the choice of $c$
guarantees that no action outside $A_t$ can take the lead at time $t+1$, since if $\bw\not\in A_t$, then
\[
(\bw-\bV_t)^\top \bH_t \ge \left|(\bw-\bV_t)^\top \bh_t\right| 
\]
so $(\bw-\bV_t)^\top \bH_{t+1}\ge 0$ and $\bw$ cannot be the new leader. It follows that we can upper bound the probability of switching as
\[
\PPt{\bV_{t+1}\neq \bV_t} \le \PPt{|A_t|>1},
\]
which leaves us with the problem of upper bounding $\PPt{|A_t|>1}$.
Similarly to the proof of Lemma~\ref{lem:leadpack}, we start analyzing $\PPt{|A_t|=1}$:
\begin{equation}\label{eq:lead1}
\begin{split}
\PPt{|A_t|=1} =& \sum_{\bv\in\Sw} \PPt{\forall \bw\neq \bv: (\bw-\bv)^\top \bH_t \ge \left\|\bw-\bv\right\|_1c}
\\
=& \sum_{\bv\in\Sw} \int\limits_{y\in\real} f_{\bv}(y)  
\PPct{\forall \bw\neq \bv: \bw^\top \bH_t \ge y+\left\|\bw-\bv\right\|_1c}{\bv^\top \bH_t = y}\,dy,
\end{split}
\end{equation}
where $f_{\bv}$ is the distribution of $\bv^\top \bH_t$.
Next we crucially use the fact that the conditional distributions of correlated Gaussian random variables are also Gaussian. In particular, defining $k(\bw,\bv) = (m-\|\bw-\bv\|_1)$, the covariances are given as
\[
\mathop{cov}\left(\bw^\top \bH_t,\bv^\top \bH_t\right) = \eta^2(m-\|\bw-\bv\|_1)t = \eta^2k(\bw,\bv)t.
\] 
Let us organize all actions $\bw\in\Sw\setminus v$ into a matrix $\bW=(\bw_1,\bw_2,\dots,\bw_{N-1})$. The conditional distribution of $\bW^\top \bH_t$ is an 
$(N-1)$-variate Gaussian distribution with mean 
\[
\mu_{\bv}(y)=\left(\bw_1^\top \bL_{t-1} + y\frac {k(\bw_1,\bv)}{m}, \bw_2^\top \bL_{t-1} + y\frac {k(\bw_2,\bv)}{m}, \dots, \bw_{N-1}^\top \bL_{t-1} + y\frac {k(\bw_{N-1},\bv)}{m}\right)^\top
\]
and covariance matrix $\Sigma_{\bv}$, given that $\bv^\top \bH_t=y$. Defining $\bK=\left(k(\bw_1,\bv),\dots,k(\bw_{N-1},\bv)\right)^\top$ and using the notation $\varphi(x) = \frac{1}{\sqrt{(2\pi)^{N-1} \left|\Sigma_v\right|}}\exp(-\frac{x^2}{2})$, we get that
\[
\begin{split}
&\PPct{\forall \bw\neq \bv: \bw^\top \bH_t \ge y+\left\|\bw-\bv\right\|_1c}{\bv^\top \bH_t = y}
\\
&\qquad\qquad=
\idotsint\limits_{z_i=y+(m-k(\bw_i,\bv))c}^\infty \phi\left(\sqrt{(\bz-\mu_{\bv}(y))^\top\Sigma_y^{-1}(\bz-\mu_{\bv}(y))}\right)\,d\bz
\\
&\qquad\qquad=
\idotsint\limits_{z_i=y+(m-k(\bw_i,\bv))c + k(\bw_i,\bv)c}^\infty \phi\left(\sqrt{\left(\bz-\mu_{\bv}(y)-c\bK\right)^\top\Sigma_y^{-1}\left(\bz-\mu_{\bv}(y)-c\bK\right)}\right) \,d\bz
\\
&\qquad\qquad=
\idotsint\limits_{z_i=y+mc}^\infty \phi\left(\sqrt{\left(\bz-\mu_{\bv}(y+mc)\right)^\top\Sigma_y^{-1}\left(\bz-\mu_{\bv}(y+mc)\right)}\right)\,d\bz
\\
&\qquad\qquad=\PPcct{\forall \bw\neq \bv: \bw^\top \bH_t \ge y+mc}{\bv^\top \bH_t = y+mc},
\end{split}
\]
where we used $\mu_{y+mc} = \mu_y + c\bK$.
Using this, we rewrite \eqref{eq:lead1} as 
\[
\begin{split}
\PPt{|A_t|=1} 
=& \sum_{\bv\in\Sw} \int\limits_{y\in \real} f_{\bv}(y)  
\PPcct{\forall \bw\neq \bv: \bw^\top \bH_t \ge y}{\bv^\top \bH_t = y}\,dy
\\
&- \sum_{\bv\in\Sw} \int\limits_{y\in\real} \bigl(f_{\bv}  (y) - f_{\bv}  (y-mc)\bigr) \PPcct{\forall \bw\neq \bv: \bw^\top \bH_t \ge y}{\bv^\top \bH_t = y}\,dy
\\
=& 1
- \sum_{\bv\in\Sw} \int\limits_{y\in\real} \bigl(f_{\bv}  (y) - f_{\bv}  (y-mc)\bigr) \PPcct{\forall \bw\neq \bv: \bw^\top \bH_t \ge y}{\bv^\top \bH_t = y}\, dy.
\end{split}
\]
To treat the remaining term, we use that $\bv^\top \bH_t$ is Gaussian with mean $\bv^\top \bL_{t-1}$ and standard deviation $\eta\sqrt{mt}$ and obtain
\[
\begin{split}
f_{\bv}  (y) - f_{\bv}  (y-mc) =&
f_{\bv}  (y) \left(1- \frac{f_{\bv}  (y-mc)}{f_{\bv}  (y)}\right)
\\
\le & f_{\bv}  (y) \left(\frac{mc^2}{2\eta^2t} - \frac{c(y-\bv^\top \bL_{t-1})}{\eta^2t}\right).
\end{split}
\]
Thus,
\[
\begin{split}
\PPt{|A_t|>1}&\le\sum_{\bv\in\Sw} \int\limits_{y\in\real} \bigl(f_{\bv}  (y) - f_{\bv}  (y-mc)\bigr) \PPcct{\forall \bw\neq \bv: \bw^\top \bH_t \ge y}{\bv^\top \bH_t = y}\,dy
\\
&\le\sum_{\bv\in\Sw} \int\limits_{y\in\real} f_{\bv}  (y) \left(\frac{mc^2}{2\eta^2t} - \frac{c(y-\bv^\top \bL_{t-1})}{\eta^2t}\right)
\PPcct{\forall \bw\neq \bv: \bw^\top \bH_t \ge y}{\bv^\top \bH_t = y}\,dy
\\
&=  \frac{mc^2}{2\eta^2t} - \frac{c\EE{\bV_t^\top \bZ_t}}{\eta^2t} \le \frac{mc^2}{2\eta^2t} + \frac{mc\EE{\left\|\bZ_t\right\|_\infty}}{\eta^2t}
\\
&= \frac{m\left\|\bh_t\right\|_\infty^2}{2\eta^2t} + \frac{m\left\|\bh_t\right\|_\infty\sqrt{2\log d}}{\eta\sqrt{t}},
\end{split}
\]
where we used the definition of  $c$ and $\EE{\left\|\bZ_t\right\|_\infty} \le \eta\sqrt{2t\log d}$ in the last step.
\end{IEEEproof}

\bibliography{ngbib,allbib,predbook}

\begin{thebibliography}{10}

\bibitem{CBLu06:Book}
N.~Cesa-Bianchi and G.~Lugosi, {\em Prediction, Learning, and Games}.
\newblock New York, NY, USA: Cambridge University Press, 2006.

\bibitem{LW94}
N.~Littlestone and M.~Warmuth, ``The weighted majority algorithm,'' {\em
  Information and Computation}, vol.~108, pp.~212--261, 1994.

\bibitem{Vov90}
V.~Vovk, ``Aggregating strategies,'' in {\em Proceedings of the third annual
  workshop on Computational learning theory (COLT)}, pp.~371--386, 1990.

\bibitem{FrSc97}
Y.~Freund and R.~Schapire, ``A decision-theoretic generalization of on-line
  learning and an application to boosting,'' {\em Journal of Computer and
  System Sciences}, vol.~55, pp.~119--139, 1997.

\bibitem{Han57}
J.~Hannan, ``Approximation to {B}ayes risk in repeated play,'' {\em
  Contributions to the theory of games}, vol.~3, pp.~97--139, 1957.

\bibitem{Kalai03efficient}
A.~Kalai and S.~Vempala, ``Efficient algorithms for the online decision
  problem,'' in {\em Proceedings of the 16th Annual Conference on Learning
  Theory and the 7th Kernel Workshop, COLT-Kernel 2003} (B.~Sch\"olkopf and
  M.~Warmuth, eds.), (New York, USA), pp.~26--40, Springer, Aug. 2003.

\bibitem{KV05}
A.~Kalai and S.~Vempala, ``Efficient algorithms for online decision problems,''
  {\em Journal of Computer and System Sciences}, vol.~71, pp.~291--307, 2005.

\bibitem{HuPo04}
M.~Hutter and J.~Poland, ``Prediction with expert advice by following the
  perturbed leader for general weights,'' in {\em Algorithmic Learning Theory},
  pp.~279--293, Springer, 2004.

\bibitem{Pol05}
J.~Poland, ``{FPL} analysis for adaptive bandits,'' in {\em In 3rd Symposium on
  Stochastic Algorithms, Foundations and Applications (SAGA'05)}, pp.~58--69,
  2005.

\bibitem{geulen10buffering}
S.~Geulen, B.~Voecking, and M.~Winkler, ``Regret minimization for online
  buffering problems using the weighted majority algorithm,'' in {\em
  Proceedings of the Twenty-Third Conference on Computational Learning Theory},
  pp.~132--143, 2010.

\bibitem{gyorgy11nearoptimal}
A.~Gy\"orgy and G.~Neu, ``Near-optimal rates for limited-delay universal lossy
  source coding,'' in {\em Proceedings of the {IEEE} International Symposium on
  Information Theory ({ISIT})}, pp.~2344--2348, 2011.

\bibitem{even-dar09OnlineMDP}
E.~Even-Dar, S.~Kakade, and Y.~Mansour, ``Online {M}arkov decision processes,''
  {\em Mathematics of Operations Research}, vol.~34, no.~3, pp.~726--736, 2009.

\bibitem{neu10o-mdp}
G.~Neu, A.~Gy\"orgy, {\text{Cs}}.~Szepesv\'ari, and A.~Antos, ``Online {M}arkov
  decision processes under bandit feedback,'' in {\em Advances in Neural
  Information Processing Systems 23}, 2010.

\bibitem{Fel68}
W.~Feller, {\em An Introduction to Probability Theory and its Applications,
  Vol.\ 1}.
\newblock New York: John Wiley, 1968.

\bibitem{GW98}
C.~Gentile and M.~Warmuth, ``Linear hinge loss and average margin,'' in {\em
  Advances in Neural Information Processing Systems (NIPS)}, pp.~225--231,
  1998.

\bibitem{KW01}
J.~Kivinen and M.~Warmuth, ``Relative loss bounds for multidimensional
  regression problems,'' {\em Machine Learning}, vol.~45, pp.~301--329, 2001.

\bibitem{GLS01}
A.~Grove, N.~Littlestone, and D.~Schuurmans, ``General convergence results for
  linear discriminant updates,'' {\em Machine Learning}, vol.~43, pp.~173--210,
  2001.

\bibitem{TW03}
E.~Takimoto and M.~Warmuth, ``Paths kernels and multiplicative updates,'' {\em
  Journal of Machine Learning Research}, vol.~4, pp.~773--818, 2003.

\bibitem{WK08}
M.~Warmuth and D.~Kuzmin, ``Randomized online pca algorithms with regret bounds
  that are logarithmic in the dimension,'' {\em Journal of Machine Learning
  Research}, vol.~9, pp.~2287--2320, 2008.

\bibitem{HW09}
D.~P. Helmbold and M.~Warmuth, ``Learning permutations with exponential
  weights,'' {\em Journal of Machine Learning Research}, vol.~10,
  pp.~1705--1736, 2009.

\bibitem{HKW10}
E.~Hazan, S.~Kale, and M.~Warmuth, ``Learning rotations with little regret,''
  in {\em Proceedings of the 23rd Annual Conference on Learning Theory (COLT)},
  pp.~144--154, 2010.

\bibitem{KWK10}
W.~Koolen, M.~Warmuth, and J.~Kivinen, ``Hedging structured concepts,'' in {\em
  Proceedings of the 23rd Annual Conference on Learning Theory (COLT)},
  pp.~93--105, 2010.

\bibitem{CL12}
N.~Cesa-Bianchi and G.~Lugosi, ``Combinatorial bandits,'' {\em Journal of
  Computer and System Sciences}, vol.~78, pp.~1404--1422, 2012.

\bibitem{audibert2011minimax}
J.~Y. Audibert, S.~Bubeck, and G.~Lugosi, ``Minimax policies for combinatorial
  prediction games,'' in {\em Conference on Learning Theory}, 2011.

\bibitem{audibert12regret}
J.~Y. Audibert, S.~Bubeck, and G.~Lugosi, ``Regret in online combinatorial
  optimization,'' {\em Manuscript}, 2012.

\bibitem{BoLuMa13}
S.~Boucheron, G.~Lugosi, and P.~Massart, {\em Concentration inequalities:A
  Nonasymptotic Theory of Independence}.
\newblock Oxford University Press, 2013.

\end{thebibliography}

\appendix
 \begin{IEEEproof}[Proof of the first statement of Theorem~\ref{thm:combbound}]
The proof is based on the proof of Theorem~4.2 of \cite{CBLu06:Book} and Theorem~3 of \cite{audibert12regret}. The main difference from those proofs is that the standard deviation of our perturbations changes over time, however, this issue is very easy to treat.
First, we define an infeasible ``forecaster'' that peeks one step into the future and uses perturbation $\hZ_t = \sqrt{t} \bX_1$:
 \[
 \hV_t = \arg\min_{w\in\Sw} \bw^\top \left(\bL_t+\hZ_t\right).
 \]
 Using Lemma~3.1 of \cite{CBLu06:Book}, we get
 \[
 \sum_{t=1}^{n} \hV_t^\top (\bloss_{t} + (\hZ_t - \hZ_{t-1})) \le \bv^\top (\bL_n +  \hZ_{n}).
 \]
 After reordering, we obtain
 \[
 \begin{split}
 \sum_{t=1}^n \bV_t^\top \bloss_t &\le \bv^\top \bL_n +  \bv^\top \hZ_{n} + \sum_{t=1}^{n} (\bV_{t}-\hV_{t})^\top \bloss_{t} - \sum_{t=1}^{n} \hV_t^\top (\hZ_t - \hZ_{t-1})
 \\
 &= \bv^\top \bL_n +  \bv^\top \hZ_{n} + \sum_{t=1}^{n} (\bV_{t}-\hV_{t})^\top \bloss_{t} +  \sum_{t=1}^{n} (\sqrt{t-1}-\sqrt{t}) \hV_t^\top \bX_1
 \end{split}
 \]
 The last term can be bounded as
 \[
 \begin{split}
 \sum_{t=1}^{n} (\sqrt{t-1}-\sqrt{t}) \hV_t^\top \bX_1 
 \le& \sum_{t=1}^{n} (\sqrt{t}-\sqrt{t-1}) \left|\hV_t^\top \bX_1\right|
 \\
 \le& m \sum_{t=1}^{n} (\sqrt{t}-\sqrt{t-1}) \left\|\bX_1\right\|_\infty 
 \\
 \le& m \sqrt{n} \left\|\bX_1\right\|_\infty.
 \end{split}
 \]
 Taking expectations, we obtain the bound
 \[
 \EE{\wh{L}_n} - \bv^\top \bL_n \le \sum_{t=1}^n \EE{(\bV_{t}-\hV_{t})^\top \bloss_{t}} + \eta m \sqrt{2n\log d},
 \]
 where we used $\EE{\left\|\bX_1\right\|_\infty}\le\eta \sqrt{2\log d}$. That is, we are left with the problem of bounding $\EE{(\bV_{t}-\hV_{t})^\top \bloss_{t}}$ for each $t\ge 1$.

 To this end, let
 \[
 \bv(\bz) = \arg\min_{\bw\in\Sw} \bw^\top \bz
 \]
 for all $\bz\in\real^d$, and also 
 \[
 F_t(\bz) = \bv(\bz)^\top \bloss_t\,.
 \]
 Further, let $f_t(\bz)$ be the density of $\bZ_t$, which coincides with the density of $\hZ_t$. 
 We have
 \[
 \begin{split}
 \EE{\bV_{t}^\top \bloss_t} =& 
 \EE{F_t(\bL_{t-1}+\bZ_{t})}
 \\
 =&
 \int_{\bz\in \real^d} f_t(\bz) F_t(\bL_{t-1} + \bz)\, d\bz
 \\
 =& \int_{\bz\in \real^d} f_t(\bz) F_t(\bL_{t} - \bloss_t + \bz)\, d\bz
 \\
 =& \int_{\bz\in \real^d} f_t(\bz + \bloss_t) F_t(\bL_{t} + \bz)\, d\bz
 \\
 =& \EE{F_t(\bL_{t}+\hZ_{t})} + \int_{\bz\in \real^d} \left(f_t(\bz+\bloss_t) - f_t(\bz)\right) F(\bL_{t} + \bz)\, d\bz
 \\
 =& \EE{\hV_{t}^\top \bloss_t} + \int_{\bz\in \real^d} \left(f_t(\bz) - f_t(\bz -\bloss_t)\right) F(\bL_{t-1} + \bz)\, d\bz\,.
 \end{split}
 \]
 The last term can be upper bounded as
 \[
 \begin{split}
 &\int_{\bz\in \real^d} f_t(\bz)\left(1-\exp\left(\frac{(\bz-\bloss_t)^\top \bloss_t}{\eta^2 t}\right)\right) F_t(\bL_{t-1} + \bz)\, d\bz
 \\
 &\qquad\qquad\le -\int_{\bz\in \real^d} f_t(\bz)\left(\frac{(\bz-\bloss_t)^\top \bloss_t}{\eta^2 t}\right) F(\bL_{t-1} + \bz) \, d\bz
 \\
 &\qquad\qquad\le \frac{\EE{\bV_t^\top \bloss_t} \left\|\bloss_t\right\|_2^2}{\eta^2 t} + \frac{m }{\eta^2 t}\int_{\bz\in \real^d} f_t(\bz)\left|\bz^\top \bloss_t\right|  \, d\bz
 \\
 &\qquad\qquad\le \frac{md}{\eta^2 t} + \frac{m}{\eta^2 t} \int_{\bz\in \real^d} f_t(\bz)\left\|\bz\right\|_1  \, d\bz
 \\
 &\qquad\qquad= \frac{md}{\eta^2 t} + \sqrt{\frac{2}{\pi}}\cdot\frac{m d}{\eta \sqrt{t}}\,,
 \end{split}
 \]
where we used $\EE{\left\|\bZ_t\right\|_1} = \eta d\sqrt{2t/\pi}$ in the last step.
Putting everything together, we obtain the statement of the theorem as
 \[
 \begin{split}
 \EE{\wh{L}_n} - \bv^\top \bL_n &\le \sum_{t=1}^n\frac{md}{\eta^2 t} + \sum_{t=1}^n \sqrt{\frac{2}{\pi}}\cdot\frac{m d}{\eta \sqrt{t}} + \eta m \sqrt{2t\log d}
 \\
 &\le \frac{2m d\sqrt{n}}{\eta} + \eta m \sqrt{2n\log d} + \frac{md(\log n + 1)}{\eta^2}\,.
 \end{split}
 \]
\end{IEEEproof}
\end{document}